\documentclass[runningheads]{llncs}
\usepackage[T1]{fontenc}
\usepackage{enumitem}
\usepackage{graphicx}
\usepackage{booktabs} 
\usepackage{array}
\newcolumntype{P}[1]{>{\raggedright\arraybackslash}p{#1}}

\begin{document}
%

\title{Pretrained, Curriculum-Tuned, and Ensembled: A Tracer-Aware Interactive Segmentation Pipeline for AutoPET V}
\titlerunning{A Tracer-Aware Interactive Segmentation Pipeline for AutoPET V}

%
%
\author{
Xinglong Liang\inst{1,2}$^{\dagger}$\orcidID{0009-0001-3813-6726}
\and
Chunyao Lu\inst{1,2}$^{\dagger}$\orcidID{0000-0002-6911-4036}
\and
Tianyu Zhang\inst{1,2}\orcidID{0000-0001-9891-6874}
\and
Jiaju Huang\inst{5}\orcidID{0009-0002-8979-0031}
\and
Tao Tan\inst{5}\orcidID{0000-0002-4014-3458}
\and
Yunchao Yin\inst{3}\orcidID{0000-0002-5790-484X}
\and
Lishan Cai\inst{1,3,4}$^{\star}$\orcidID{0000-0002-3282-291X}
}

\authorrunning{X. Liang et al.}

\institute{
Department of Radiology, The Netherlands Cancer Institute,
Plesmanlaan 121, 1066 CX, Amsterdam, the Netherlands
\and
Department of Radiology and Nuclear Medicine,
Radboud University Medical Centre, Geert Grooteplein 10,
6525 GA, Nijmegen, The Netherlands
\and
Department of Radiation Oncology, Amsterdam University Medical Center,
De Boelelaan 1117, 1081 HV Amsterdam, the Netherlands
\and
GROW School for Oncology and Developmental Biology,
Maastricht University Medical Centre, P. Debyelaan 25,
6229 AZ, Maastricht, The Netherlands
\and
Faculty of Applied Sciences, Macao Polytechnic University,
999078, Macao Special Administrative Region of China
}

\maketitle

\let\thefootnote\relax
\footnotetext{
$^{\dagger}$ These authors contributed equally.
$^{\star}$ Corresponding author:
\email{l.cai@amsterdamumc.nl }
}
\begin{abstract}
Interactive lesion segmentation in whole-body PET/CT requires a model to provide a strong initial prediction while also responding efficiently to sparse corrective scribbles during inference. This setting is particularly challenging because tracer distributions, physiological uptake patterns, lesion appearance, and acquisition characteristics differ substantially between FDG and PSMA studies. We present \textbf{TRIAGE}, Tracer-aware Refinement via Interactive Anatomy-Guided sEgmentation. The core segmentation backbone is a 3D STU-Net initialized through masked autoencoding pre-training with an asynchronous masking strategy, aiming to learn transferable anatomical and cross-modal representations from PET/CT volumes before task-specific fine-tuning. In parallel, we train an auxiliary organ segmentation model whose predictions provide explicit anatomical context and help distinguish physiological uptake from malignant lesions. A dedicated tracer classifier first routes each study to an FDG- or PSMA-specific branch. Within each branch, a first-stage segmentation model consumes CT, PET, and organ context to generate an initial lesion mask. The initial prediction is then combined with cumulative foreground/background scribbles and refined by a second interactive segmentation network. The FDG and PSMA branches share the same overall processing pipeline but are trained independently to account for tracer-specific appearance and error modes. We additionally employ curriculum-style training and model ensembling to improve robustness across interaction steps and heterogeneous cohorts. Experiments are conducted using the official AutoPET V data and ten-fold split; quantitative results, ablations, and final test-set performance are left as placeholders to be completed after the challenge evaluation. Our framework is designed to jointly optimize strong automatic initialization and rapid correction under sparse human interaction. Code: https://github.com/Liiiii2101/AUTOPET2026-MEDAI.
\end{abstract}

\section{Introduction}

Positron Emission Tomography and Computed Tomography (PET-CT) is widely used in oncology by combining functional information from PET with anatomical information from CT~\cite{groheux2013performance,endo2006pet}. Accurate lesion segmentation from whole-body PET-CT is important for quantitative tumor burden assessment and treatment response evaluation. However, automated segmentation remains challenging because lesions can be small, spatially scattered, and highly heterogeneous in appearance. In addition, physiological tracer uptake in normal organs can resemble pathological uptake and lead to false-positive predictions~\cite{barrington2003limitations,gontier2008high}.

The AutoPET V challenge focuses on \emph{interactive} lesion segmentation in whole-body PET-CT. Instead of producing only a single automatic prediction, the segmentation model is expected to generate an initial lesion mask and subsequently refine it according to sparse user-provided scribbles. This setting introduces additional requirements beyond conventional automatic segmentation: the model should provide a reliable initial prediction, effectively interpret positive and negative corrections, and progressively improve the segmentation with each interaction. Moreover, the challenge includes both FDG and PSMA PET-CT scans, whose physiological uptake distributions and lesion appearances can differ considerably. Therefore, robust segmentation requires both multimodal PET-CT information and tracer-specific modeling.

Recent medical image segmentation frameworks such as nnU-Net~\cite{isensee2021nnu} and STU-Net~\cite{huang2023stu} have demonstrated strong performance on volumetric segmentation tasks. PET-CT-specific studies have further shown the benefit of combining metabolic PET information with anatomical CT context~\cite{zhao2019tumor,fu2021multimodal,liang2026leveraging}. Nevertheless, directly training a single segmentation model from limited annotated PET-CT data may not fully exploit the complementary information between PET and CT, while physiological uptake can still introduce substantial false positives. These observations motivate the use of pre-training, anatomical priors, and tracer-specific models for the AutoPET V setting.

In this work, we present a tracer-aware interactive segmentation pipeline for AutoPET. Our framework is based on STU-Net-Small and consists of a pre-training stage followed by a two-stage interactive segmentation pipeline. We first perform masked autoencoder\cite{he2022masked} pre-training using PET and CT with independently sampled spatial masks. For downstream segmentation, FDG and PSMA scans are processed by separately trained models following the same pipeline. The first-stage network takes PET and CT as input and generates an initial lesion segmentation. The second-stage network further incorporates anatomical organ masks and interaction information to refine the prediction after scribble corrections. In addition, curriculum-based training and model ensembling are employed to improve robustness during interactive inference.

Our approach is designed specifically for the AutoPET V task by combining multimodal pre-training, tracer-aware specialization, anatomical guidance, and interaction-driven refinement within a unified segmentation framework. Our contributions are summarized as follows:
\begin{itemize}[leftmargin=*]
    \item We combine MAE-pretrained STU-Net features with organ-level anatomical context to strengthen the initial lesion segmentation and suppress physiological false positives.
    \item We formulate interactive refinement as a second-stage segmentation problem conditioned on the image channels, initial prediction, and cumulative scribble guidance.
    \item We use curriculum-style training over interaction difficulty and ensemble multiple folds to improve robustness across centers, tracers, and correction steps.
\end{itemize}

\section{Methods}
\begin{figure*}[t]
\centerline{\includegraphics[width=\textwidth]{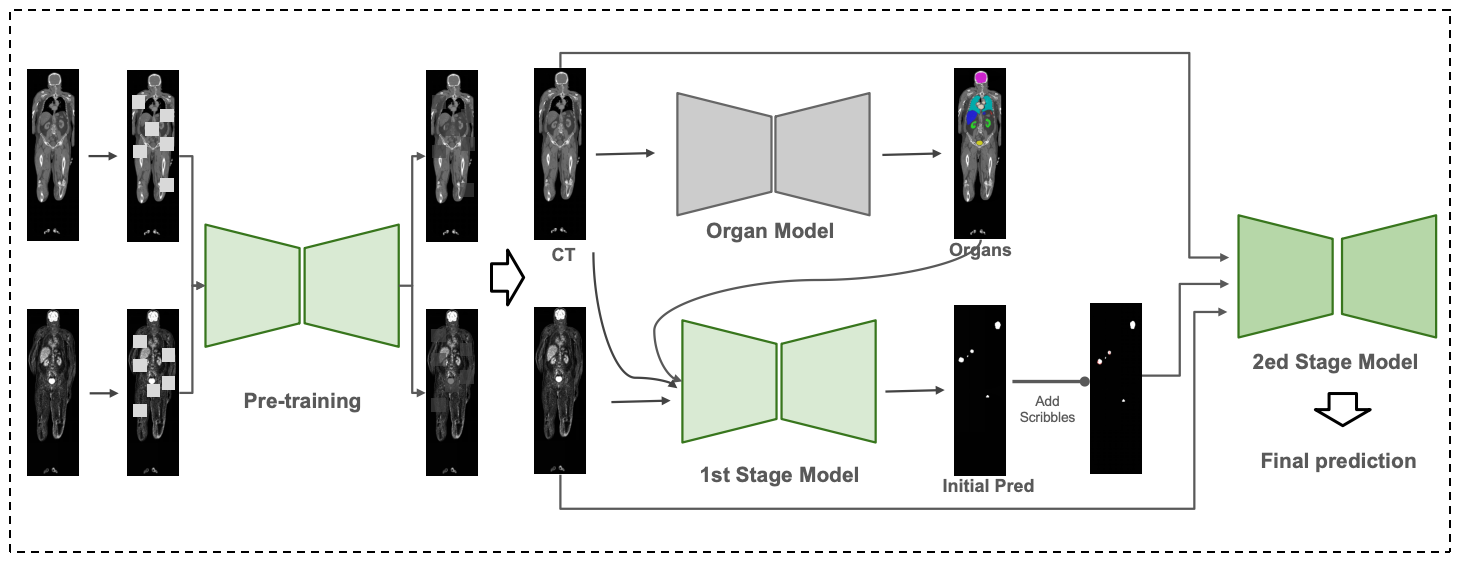}}
\caption{Overview of the proposed TRIAGE framework.
A STU-Net backbone is first pre-trained on paired PET/CT volumes using masked autoencoding. For downstream segmentation, a tracer classifier routes each case to an FDG- or PSMA-specific pipeline. A seperate organ model provides anatomical context, and the first-stage model combines PET, CT, and organ predictions to generate the initial lesion prediction. The second-stage model then integrates the image channels, organ predictions, initial mask, and cumulative foreground/background scribbles to iteratively refine the segmentation.}
\label{fig:pipeline}
\end{figure*}

Our framework consists of a tracer-aware pre-training and interactive segmentation pipeline. The overall workflow is illustrated in Fig.~\ref{fig:pipeline}. We first pre-train a STU-Net-Small backbone using a masked autoencoder strategy on paired PET and CT volumes. To better exploit the complementary information between the two modalities, PET and CT are masked independently using different spatial mask locations, with a masking ratio of 0.5 for each modality.

For downstream segmentation, FDG and PSMA cases are processed by separately trained models with the same network architecture and training pipeline. A tracer classification network is used to determine whether an input case belongs to the FDG or PSMA domain and route it to the corresponding segmentation model.

The segmentation pipeline contains two stages. The first-stage model takes PET, CT, and anatomical organ masks as input and generates an initial lesion prediction. The second-stage model further refines this prediction by incorporating PET, CT, anatomical organ masks, the initial prediction, and user-provided scribble information. Curriculum-based training is adopted for the refinement stage, and multiple trained models are ensembled during inference to improve robustness.

\subsection{Data}
We used only the official AutoPET V training dataset and did not incorporate any external imaging data. The dataset contains 1,014 FDG PET/CT studies from 900 patients and 597 PSMA PET/CT studies from 378 patients. The FDG cohort was acquired at the University Hospital Tübingen and includes patients with malignant melanoma, lymphoma, or lung cancer, as well as negative control cases. The PSMA cohort was acquired at LMU University Hospital Munich and contains prostate cancer studies both with and without PSMA-avid tumor lesions.

Each study consists of a whole-body CT volume, a corresponding PET volume in SUV, and a binary manual segmentation of tracer-avid tumor lesions. PET and CT were acquired in the same imaging session and were provided in a spatially aligned form. The released training data follow the nnU-Net NIfTI format, with CT and PET represented as separate image channels. Tumor lesions were manually annotated by experienced readers based on joint PET/CT assessment and clinical information.

We followed the nnU-Net framwork for model development. FDG and PSMA cases were treated as two separate tracer domains, and independent models were trained for the two cohorts using the same overall pipeline. 

\subsection{Data pre-processing}
We used the default nnU-Net pre-processing and planning pipeline. For self-supervised pre-training, PET/CT volumes from both tracers were sampled using fixed $128\times128\times128$ voxel patches. For supervised segmentation, tracer-specific nnU-Net configurations were used. FDG images were resampled to a target spacing of $(3.0, 2.03, 2.03)$ mm and trained with $128\times128\times128$ patches. PSMA images were resampled to $(4.07, 3.27, 4.07)$ mm and trained with $112\times192\times112$ patches. All values were automatically determined by the default nnU-Net planning procedure.

\subsection{Data post-processing}
To suppress false positives arising from physiological tracer uptake, we applied a tracer-specific SUV thresholding step to the predicted segmentation. For each voxel predicted as lesion, the corresponding PET SUV value was checked against a fixed threshold, with voxels below the threshold reassigned to background; the threshold was set to 1.5 g/mL for FDG and 1.0 g/mL for PSMA to reflect the differing physiological uptake characteristics of the two tracers. The thresholded prediction was then binarized to produce the final lesion mask.

\subsection{Training and test parameters}
Using the pretrained encoder, segmentation training proceeded in three stages. First, the encoder was frozen and the decoder was warmed up alone for 50 epochs with a small learning rate of $1\times10^{-5}$. Next, the encoder was unfrozen and the entire network was warmed up for a further 50 epochs. Finally, the full model was trained with a learning rate of $1\times10^{-4}$ for a total of 1000 epochs following the default nnU-Net training strategy~\cite{isensee2021nnu}, optimizing the sum of the Dice and cross-entropy losses. More details can be found in Table~\ref{tab1}.

\subsection{Github repository}
Link to Github repository: https://github.com/Liiiii2101/AUTOPET2026-MEDAI

\section{Results}
We evaluated the proposed framework using ten-fold cross-validation separately on the FDG and PSMA cohorts. Tables~\ref{tab:stage1_dice} and~\ref{tab:stage2_dice} summarize the validation Dice scores obtained by the first-stage automatic segmentation model and the second-stage refinement model, respectively.

For the first-stage model, the mean validation Dice was $0.5889 \pm 0.0537$ for FDG and $0.5833 \pm 0.0293$ for PSMA. Performance varied across folds, with FDG Dice scores ranging from 0.5093 to 0.6551 and PSMA scores ranging from 0.5343 to 0.6234. These results reflect the difficulty of generating a reliable initial whole-body lesion segmentation from PET/CT alone, particularly in the presence of heterogeneous lesion appearance and physiological tracer uptake.

The second-stage model consistently achieved higher average Dice scores than the first-stage model. The mean Dice increased to $0.6175 \pm 0.0433$ for FDG and $0.6405 \pm 0.0270$ for PSMA, corresponding to absolute improvements of 0.0286 and 0.0572, respectively. The improvement was particularly pronounced for PSMA, suggesting that the additional anatomical context, initial prediction, and interaction guidance used by the refinement stage provide complementary information beyond the original PET/CT channels.

Across the evaluated folds, the second-stage PSMA model achieved a maximum Dice of 0.6853, while the FDG model reached a maximum Dice of 0.6766. For FDG, fold 4 training was incomplete and was therefore excluded from the reported mean and standard deviation; the remaining nine folds were used for the aggregate statistics.

Overall, these cross-validation results demonstrate that the second-stage refinement strategy improves segmentation accuracy over the initial automatic prediction for both tracer domains. The larger improvement observed for PSMA further supports the use of tracer-specific models rather than a single shared model across both PET tracers.

\begin{table}[t]
\centering
\caption{Stage-1 validation Dice scores for FDG and PSMA across 10 folds.}
\label{tab:stage1_dice}
\begin{tabular}{c|cc}
\hline
\textbf{Fold} & \textbf{FDG Dice} & \textbf{PSMA Dice} \\
\hline
0 & 0.5534 & 0.5990 \\
1 & 0.5750 & 0.5343 \\
2 & 0.6551 & 0.5949 \\
3 & 0.5124 & 0.5712 \\
4 & 0.6464 & 0.5579 \\
5 & 0.6365 & 0.6074 \\
6 & 0.6221 & 0.5966 \\
7 & 0.5642 & 0.6234 \\
8 & 0.5093 & 0.6025 \\
9 & 0.6145 & 0.5459 \\
\hline
\textbf{Mean} & \textbf{0.5889} & \textbf{0.5833} \\
\textbf{Mean $\pm$ Std.} & \textbf{0.5889 $\pm$ 0.0537} & \textbf{0.5833 $\pm$ 0.0293} \\
\hline
\end{tabular}
\end{table}

\begin{table}[t]
\centering
\caption{Stage-2 validation Dice scores for FDG and PSMA across 10 folds. `\_' indicates that training was incomplete; parameters from the intermediate checkpoint were used.}
\label{tab:stage2_dice}
\begin{tabular}{c|cc}
\hline
\textbf{Fold} & \textbf{FDG Dice} & \textbf{PSMA Dice} \\
\hline
0 &0.5604  &0.6717  \\
1 & 0.6278 &0.6134  \\
2 & 0.6621 & 0.6383 \\
3 & 0.5565 & 0.5937 \\
4 & - &  0.6153\\
5 & 0.6393 & 0.6853 \\
6 & 0.6766 &  0.6609\\
7 & 0.6270 & 0.6340 \\
8 & 0.5630 & 0.6357 \\
9 & 0.6446 &  0.6562\\
\hline
\textbf{Mean} & \textbf{0.6175} & \textbf{0.6405} \\
\textbf{Mean $\pm$ Std.} & \textbf{0.6175 $\pm$ 0.0433} & \textbf{0.6405 $\pm$ 0.0270} \\
\hline
\end{tabular}
\end{table}

\begin{table}[t]
\centering
\caption{Ablation study of network architecture and pre-training on one FDG validation fold from an earlier five-fold experimental setting. Organ context was not used in these experiments. All models were trained and evaluated using the same data split and validation protocol.}
\label{tab:pretraining_ablation}
\begin{tabular}{l|c|c|c}
\hline
\textbf{Method} & \textbf{Parameters (M)} & \textbf{FLOPs} & \textbf{Dice} \\
\hline
nnU-Net & 30.79 & 526.29 & 0.3990 \\
STU-Net & 14.55 & 138.70 & 0.3890 \\
STU-Net + Pre-training & 14.55 & 138.70 & \textbf{0.4889} \\
\hline
\end{tabular}
\end{table}

\section{Discussion}

The cross-validation experiments show that the proposed two-stage strategy provides a clear benefit over the initial automatic segmentation for both FDG and PSMA PET/CT. While the first-stage model is responsible for generating a strong initial lesion mask without user interaction, the second-stage model has access to additional information, including anatomical organ predictions, the initial segmentation, and foreground/background scribbles. The improvement from a mean Dice of 0.5889 to 0.6175 for FDG and from 0.5833 to 0.6405 for PSMA indicates that these additional cues are useful for correcting errors that cannot be resolved from PET and CT alone.

The larger Stage-2 improvement observed for PSMA suggests that anatomical and interaction guidance may be particularly beneficial for this tracer. PSMA PET has tracer-specific physiological uptake patterns that differ substantially from FDG, motivating our decision to train independent FDG and PSMA models rather than forcing both domains into a single segmentation network. Tracer-specific training allows each model to specialize in its corresponding uptake distribution, lesion appearance, and characteristic false-positive patterns while retaining the same overall architecture and training strategy.

The ablation experiment further demonstrates the effectiveness of pre-training and the efficiency of the adopted STU-Net backbone. This experiment was conducted on a single FDG validation fold from an earlier five-fold setting without organ context, and is therefore intended only to compare the relative effects of network architecture and pre-training under the same experimental conditions. STU-Net trained from scratch achieved a mean validation Dice of 0.3890, which was comparable to the nnU-Net baseline of 0.3990, while requiring substantially fewer parameters and FLOPs (14.55M parameters and 138.70 GFLOPs vs.\ 30.79M and 526.29 GFLOPs). After initializing STU-Net with the proposed pre-trained weights, the validation Dice increased to 0.4889, corresponding to an absolute gain of 0.0999 over the same architecture trained from scratch. Importantly, this improvement was achieved without increasing the model size or inference complexity.

The current results demonstrate a consistent advantage of the refinement stage across both tracer domains and support the overall design of combining tracer-specific specialization, anatomical context, and interaction-aware refinement for whole-body PET/CT lesion segmentation.
\section{Conclusion}

In this article, we presented a tracer-aware two-stage framework for interactive whole-body lesion segmentation in FDG and PSMA PET/CT. The approach combines masked-autoencoder pre-training, tracer-specific segmentation models, anatomical organ guidance, and scribble-conditioned refinement within a unified pipeline. Ten-fold cross-validation showed that the second-stage refinement model improved the mean Dice from 0.5889 to 0.6175 for FDG and from 0.5833 to 0.6405 for PSMA compared with the initial segmentation stage. These results demonstrate the benefit of incorporating anatomical context and interaction information for refining PET/CT lesion predictions. Future work will evaluate the framework on the official AutoPET V test set and further investigate the contributions of individual components through systematic ablation studies.

\begin{table}[ht]
\caption{Algorithm details}\label{tab1}

\begin{tabular}{P{0.2\textwidth}P{0.2\textwidth}P{0.2\textwidth}P{0.2\textwidth}P{0.2\textwidth}} 
\toprule
\textbf{Team name} & \textbf{algorithm name} & \textbf{data pre-processing} & \textbf{data post-processing} & \textbf{training data augmentation} \\
\midrule
MEDAI & TRIAGE — Tracer-aware Refinement via Interactive Anatomy-Guided sEgmentation &  nnUNet Preprocessing & SUV thresholding (1.5 g/mL FDG, 1.0 g/mL PSMA) & Random Brightness, Random Gamma, Random Rotation \\
\bottomrule
\end{tabular}

\vspace{2em}

\begin{tabular}{P{0.2\textwidth}P{0.2\textwidth}P{0.2\textwidth}P{0.2\textwidth}P{0.2\textwidth}} 
\toprule
\textbf{test time augmentation} & \textbf{ensembling}& \textbf{standardized framework?}  & \textbf{network architecture} & \textbf{loss} \\
\midrule
- & Tracer-based routing (Based on the pre-trained 10-fold FDG and 10-fold PSMA models.)& nnUNet v2 (3D) & STUNet-small (3D) & DSC + CE \\
& & & & \\
\bottomrule
\end{tabular}

\vspace{2em}

\begin{tabular}{P{0.2\textwidth}P{0.2\textwidth}P{0.2\textwidth}P{0.2\textwidth}P{0.2\textwidth}} 
\toprule
\textbf{training data} & \textbf{data/model dimensionality and size}& \textbf{use of pre-trained models} & \textbf{GPU hardware for training}\\
\midrule
1014 FDG + 597 PSMA PET-CT of autoPET & For FDG, the 3D volume size is $128 \times 128 \times 128$, while for PSMA, it is $112 \times 192 \times 112$. & Own developed & 1x Nvidia a6000 \\
& & & & \\
\bottomrule
\end{tabular}
\end{table}


\begin{credits}
\subsubsection{\ackname} The authors would like to acknowledge the Research High-Performance Computing (RHPC) facility of the Netherlands Cancer Institute (NKI).

\subsubsection{\discintname}
The authors have no competing interests to declare that are relevant to the content of this article.
\end{credits}
%
%
%
\bibliographystyle{splncs04}
\bibliography{mybibliography}

\end{document}